\documentclass[letterpaper, 10 pt, conference]{ieeeconf}  

\IEEEoverridecommandlockouts                              

\usepackage{graphicx}
\usepackage{amsmath}
\usepackage{amssymb}
\usepackage{booktabs}
\usepackage{multirow}
\usepackage[caption=false,font=footnotesize]{subfig}
\usepackage{hyperref}
\usepackage[dvipsnames]{xcolor}

\title{\LARGE \bf 
DiFuse-Net: RGB and Dual-Pixel Depth Estimation using Window Bi-directional Parallax Attention and Cross-modal Transfer Learning
}

\author{Kunal Swami, Debtanu Gupta, Amrit Kumar Muduli, Chirag Jaiswal and Pankaj Kumar Bajpai
\thanks{Authors are affiliated with Visual Intelligence Team,
        Samsung Research India Bangalore. Correspondence email: {\tt\small kunal.swami@samsung.com}}%
}

\begin{document}

\maketitle
\thispagestyle{empty}
\pagestyle{empty}

\begin{abstract}
	
Depth estimation is crucial for intelligent systems, enabling applications from autonomous navigation to augmented reality. While traditional stereo and active depth sensors have limitations in cost, power, and robustness, dual-pixel (DP) technology, ubiquitous in modern cameras, offers a compelling alternative. This paper introduces DiFuse-Net, a novel modality decoupled network design for disentangled RGB and DP based depth estimation. DiFuse-Net features a window bi-directional parallax attention mechanism (WBiPAM) specifically designed to capture the subtle DP disparity cues unique to smartphone cameras with small aperture. A separate encoder extracts contextual information from the RGB image, and these features are fused to enhance depth prediction. We also propose a Cross-modal Transfer Learning (CmTL) mechanism to utilize large-scale RGB-D datasets in the literature to cope with the limitations of obtaining large-scale RGB-DP-D dataset. Our evaluation and comparison of the proposed method demonstrates its superiority over the DP and stereo-based baseline methods. Additionally, we contribute a new, high-quality, real-world RGB-DP-D training dataset, named Dual-Camera Dual-Pixel (DCDP) dataset, created using our novel symmetric stereo camera hardware setup, stereo calibration and rectification protocol, and AI stereo disparity estimation method.

\end{abstract}

\section{INTRODUCTION}
\label{sec:intro}

Robust 3D perception is a cornerstone of intelligent systems, enabling a wide range of capabilities from autonomous navigation \cite{kitti_ijrr2013,vkitti2_arxiv2020} and manipulation in robotics to augmented reality and scene understanding in consumer devices \cite{synthetic_dof_siggraph2018}.  Specifically in robotics, accurate and efficient depth estimation is crucial for tasks such as obstacle avoidance, path planning, object recognition, and grasping. Depth estimation, the process of inferring the distance to objects and surfaces in a scene, remains a challenging problem, particularly in dynamic and unstructured environments. 

Traditionally, depth is acquired through stereo cameras \cite{crestereo_cvpr2022,wavelet_synth_disp_cvpr2020}, multiple cameras \cite{riav_mvs_cvpr2023,geomvsnet_cvpr2023}, or active depth sensors (e.g., time-of-flight, structured light). However, these methods often face limitations in practical applications, including calibration complexity \cite{taxonomy_schar_ijcv_2002}, power consumption, cost, and environmental constraints \cite{nyuv2_eccv2012,kitti_ijrr2013}. Consequently, there is growing interest in monocular depth estimation, a cost-effective and power-efficient alternative. However, monocular depth estimation from a single RGB image is a fundamentally ill-posed problem; a given 2D image can correspond to infinitely many 3D scenes.  Even relative depth estimation methods \cite{redweb_cvpr2018,dp_depth_iccv2019,rank_loss_adobe_cvpr2020,midas_tpami2020} struggle to determine accurate ordinal relationships between all pixels due to the inherent lack of geometric information.

\begin{figure}[t!]
	\centering
	\includegraphics[width=0.98\linewidth]{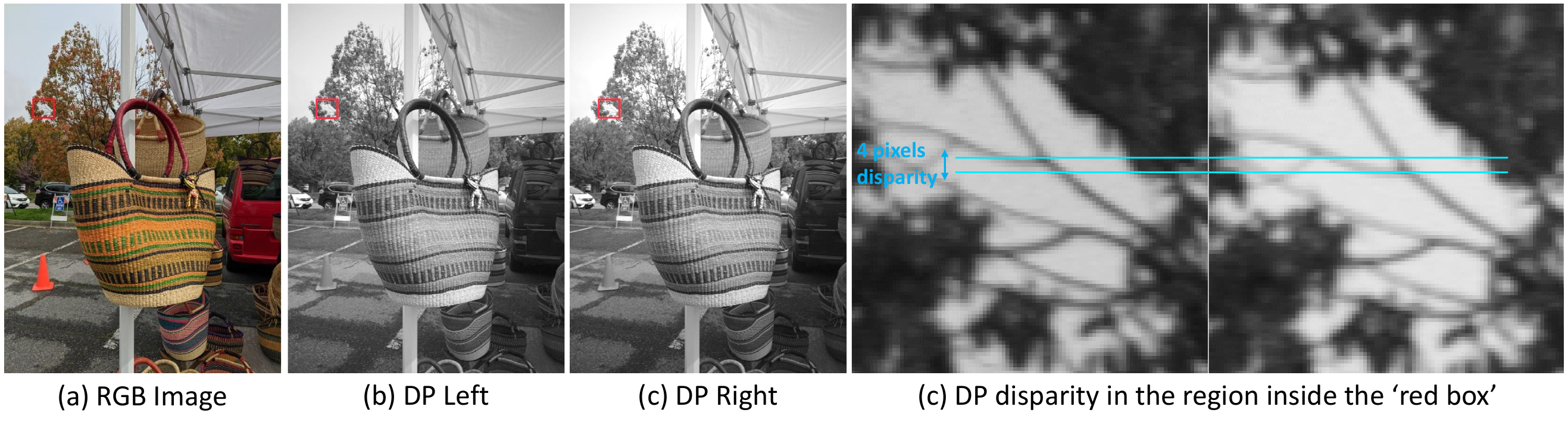}
	\vspace{-8pt}
	\caption{DP disparities from a sample captured with a Google Pixel $3$ smartphone. Note: (1). We refer to the DP images as `left' and `right' for consistency with prior DP literature, which often utilizes DSLR cameras, although the physical disparity in most smartphones is vertical due to the sensor design and arrangement. (2). The grayscale representation of DP images reflects the limitation of DP sensing to the green channel of the Bayer sensor in smartphones for cost savings.}
	\label{fig:pixel_disp}
	\vspace{-14pt}
\end{figure}

Recent advancements in camera sensors, particularly dual-pixel (DP) technology, offer a promising new avenue for single-camera depth estimation. Now ubiquitous in smartphones like the Samsung Galaxy S24, DP sensors provide two sub-images per pixel, exhibiting defocus disparities \cite{dpdnet_eccv2020, dp_depth_iccv2019}.  While originally designed for autofocus, DP sensors are also valuable for depth estimation \cite{dp_depth_iccv2019}. This makes RGB-DP based depth sensing particularly attractive for applications in mobile robotics, drones, and other embedded systems. Our objective is to develop an accurate RGB-DP based depth estimation method for smartphones, key challenges include:

\begin{enumerate}
	\itemsep0em
	\item \textbf{Subtle DP disparities}: Small smartphone apertures lead to subtle defocus disparities (see Fig.~\ref{fig:pixel_disp}), requiring specialized techniques for effective exploitation.
	\item \textbf{Lack of large-scale RGB-DP-D datasets}: Unlike RGB-D datasets \cite{midas_tpami2020, blender,hypersim_iccv2021,vkitti2_arxiv2020,nyuv2_eccv2012}, large-scale RGB-DP-D datasets require specialized capture hardware and procedures, hindering scalability and research progress.
\end{enumerate}

To address the first issue, we propose a new architecture named DiFuse-Net (\emph{di}sentangle then \emph{fuse}) for RGB-DP based depth estimation.
 
To perform disentangled and specialized processing, DiFuse-Net employs a two-branch encoder: the RGB encoder extracts global scene context, while the DP encoder leverages a novel Window Bi-directional Parallax Attention Module (WBiPAM) for effective disparity matching. A dynamic fusion module then combines information from corresponding RGB and DP encoder layers for accurate depth estimation. This design also makes DiFuse-Net robust to changes in DP disparity range across different camera apertures.

To address the second issue, we propose a Cross-modal Transfer Learning (CmTL) strategy. The modality decoupled design of DiFuse-Net presents a unique opportunity for leveraging existing large-scale RGB-D datasets to enhance depth estimation generalization and performance. The proposed CmTL approach comprises of three training stages to exploit this advantage.

Additionally, since there is no RGB-DP-D dataset with high-quality ground-truth depth, we present a new method to overcome this challenge using a symmetric stereo setup with two smartphones. Acknowledging the non-rigid nature of smartphone lens systems, we employ regular calibration and rectification protocol for each capture session. We then train an AI stereo disparity estimation method on synthetic data to generate high-quality ground-truth depth maps, ensuring the method is robust to minor rectification errors ($<3$ pixels) inevitable in our setup. Our scalable dataset capture setup and method leads to $5000$ novel training samples, and $700$ novel test samples, which is one of its kind in the literature. Fig.~\ref{fig:google_vs_our_dataset} shows samples from our new dataset named Dual-Camera Dual-Pixel (DCDP) dataset.

Following are the \textbf{major contributions} of this work:
\begin{enumerate}
	\itemsep0em
	\item A new architecture for RGB-DP based depth estimation that decouples the processing of RGB and DP images for adequate global scene understanding from the RGB image and specialized processing of the DP images to discern the defocus disparity of each pixel.
	\item A new WBiPAM based siamese encoder for processing the DP images to extract the defocus disparity information at multiple scales effectively.
	\item A dynamic fusion module to fuse the information from the RGB and DP modality feature maps effectively.
	\item CmTL mechanism to exploit the depth estimation prior from large-scale RGB-D datasets to address the problem of obtaining large-scale RGB-DP-D datasets.
	\item A new and scalable method to obtain RGB-DP-D dataset with high quality depth ground-truth using a custom symmetric stereo camera setup made with two smartphone devices. The ground-truth depth maps in our dataset are significantly dense and have better quality than the existing dataset in the literature \cite{dp_depth_iccv2019}. Our new dataset named DCDP, containing $5000$ novel training samples, and $700$ novel test samples, will support future research on RGB-DP based depth estimation.
\end{enumerate}

\section{RELATED WORK}
\label{sec:relatedwork}

\begin{figure}[t!]
	\centering
	\includegraphics[width=0.98\linewidth]{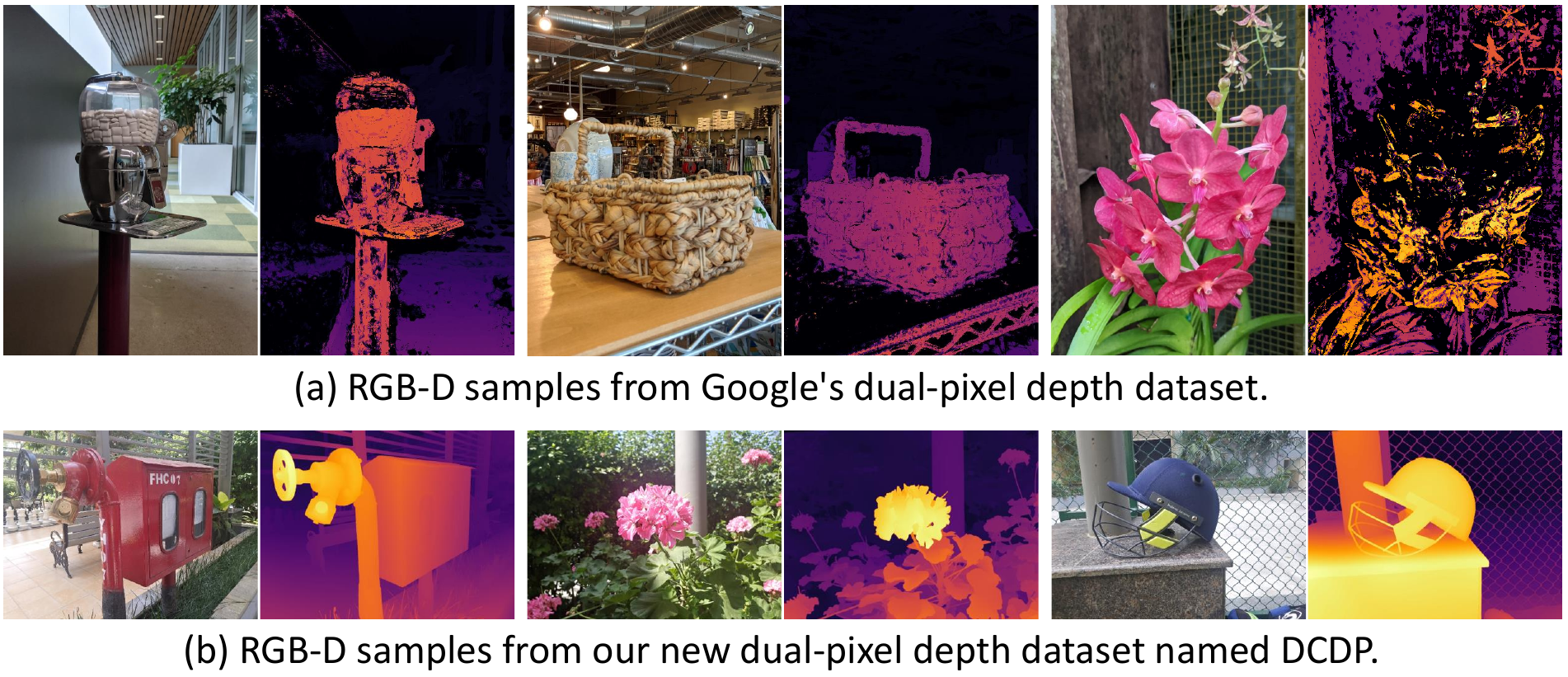}
	\vspace{-8pt}
	\caption{Qualitative difference between random RGB-D samples from our dataset vs dataset in \cite{dp_depth_iccv2019}. As visually evident, our new DCDP dataset exhibits superior ground-truth quality, particularly in terms of density, sharpness, and boundary delineation.} 
	\label{fig:google_vs_our_dataset}
	\vspace{-14pt}
\end{figure}

Garg \textit{et al.} \cite{dp_depth_iccv2019} were the first to propose an RGB-DP based depth estimation method. The authors designed a lightweight network architecture named DPNet, which takes the channel-wise concatenated RGB and DP images as input and is trained end-to-end using an affine invariant loss function. To acquire the training dataset, they created a custom camera rig using five Google Pixel $3$ smartphones, and the ground-truth depth map was computed using multi-view stereo techniques. The ground-truth depth maps in their dataset are sparse and often erroneous, as shown in Fig.~\ref{fig:google_vs_our_dataset}. Therefore, it limits the accuracy and quality of the depth estimation. Moreover, the authors do not focus on specialized processing of the DP images to compute the defocus disparity information, which makes it less effective.

Zhang \textit{et al.} \cite{du2net_eccv2020} extended the method in \cite{dp_depth_iccv2019} by additionally using dual cameras. They kept the baseline of the dual cameras orthogonal to the DP baseline to obtain complementary disparity information from the two sources for more accurate depth estimation. Pan \textit{et al.} \cite{dp_depth_blur_cvpr2021} proposed a method for DP image simulation for DSLR images and trained a multi-task model to estimate depth and perform image deblurring simultaneously. In \cite{model_defocus_dp_iccp2020}, the authors proposed a method for modeling defocus disparity in DSLR DP images, operating on patches and stitching the outputs to recover the disparity map. The authors also provide $100$ RGB-DP-Disparity triplets of simple lab scenes using a DSLR, whereas some handful RGB-DP images from Pixel smartphone were used only for qualitative evaluation. As it is pointed in \cite{model_defocus_dp_iccp2020}, it is important to note that smartphone and DSLR DPs are significantly different, therefore, model trained on DSLR dataset cannot be evaluated on smartphone dataset and vice-versa.

In this work, we focus on explicitly and effectively utilizing the DP defocus disparity cues. More specifically, we decouple the processing of the RGB and DP images to exploit the information from each modality effectively.

\newcommand{\googoursdbfigwidth}[0]{0.15\linewidth}

\section{PROPOSED METHOD}
\label{sec:proposed_method}

\begin{figure}[t!]
	\centering
	\includegraphics[width=1.0\linewidth]{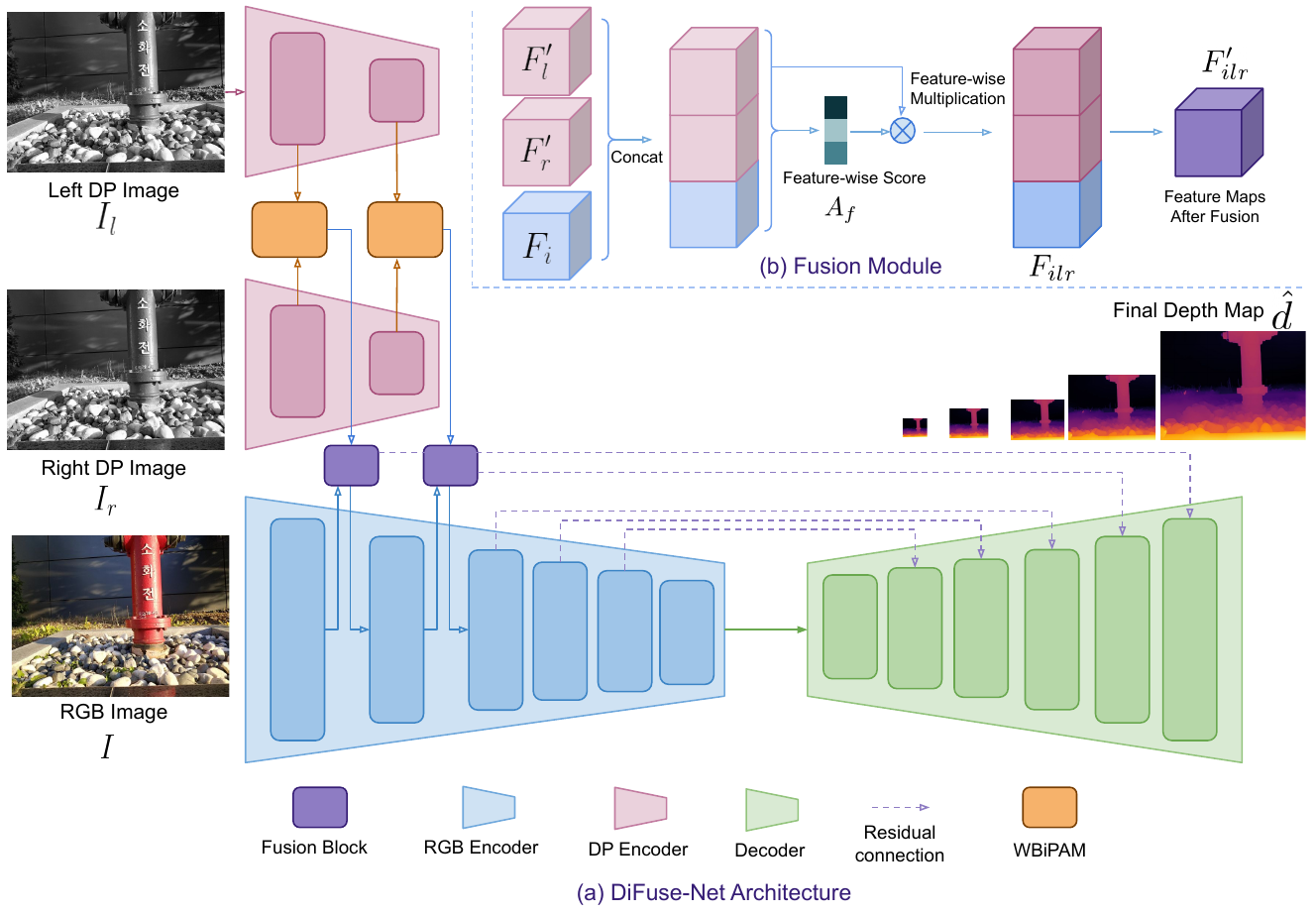}
	\vspace{-16pt}
	\caption{Architecture of the proposed DiFuse-Net model. Zoom-in using Adobe Reader.}
	\label{fig:network_arch}
	\vspace{-8pt}
\end{figure}

Fig.~\ref{fig:network_arch} shows the detailed architecture of the proposed method named DiFuse-Net. Given an RGB image $I \in \mathcal{R}^{H \times W \times 3}$, and DP image pair $I_{l}, I_{r} \in \mathcal{R}^{H \times W \times 1}$, the objective is to predict a relative depth map $\hat{d} \in \mathcal{R}^{H \times W \times 1}$ of the scene represented by $I$. Here, $H$ and $W$ represent the spatial height and width of the image, respectively.

Our investigations and ablation study reveal that a na\"ive concatenation of the RGB and DP images as input to the network results in sub-optimal performance because RGB and DP modalities contain distinct information. While RGB images provide the global context for holistic scene understanding, DP images provide localized disparity cues that assist in resolving the ordinal relationships of the pixels, leading to accurate depth prediction. This understanding motivates us to adopt an independent processing of these two modalities. However, stereo matching is known to be less reliable in textureless regions \cite{wavelet_synth_disp_cvpr2020,taxonomy_schar_ijcv_2002}, implying that it is better to rely on global contextual information from the RGB image for these regions. We intend to adaptively combine the RGB and DP modality information in the feature space, enabling the model to leverage the strengths of each modality while mitigating their respective limitations.

As shown in Fig.~\ref{fig:network_arch}, DiFuse-Net contains two distinct encoding modules: one encoder for encoding the RGB image and a siamese encoder for encoding the DP image pair. The siamese encoder contains the WBiPAM module for effective correspondence matching between the DP images. The fusion module adaptively fuses the RGB and DP modality information, which is passed to the successive blocks in the encoder. A UNet-style decoder \cite{unet_miccai2015} finally predicts the depth map. In the following subsections, we explain each of these modules of DiFuse-Net in detail.

\subsection{RGB Encoder}
\label{subsec:rgbencoder}

The RGB encoder processes the input images $I \in \mathcal{R}^{H \times W \times 3}$. It employs an EfficientNet-Lite$3$ backbone \cite{effnet_icml2019}, initialized with pre-trained ImageNet \cite{imagenet_cvpr2009} weights, for feature extraction. The network progressively downsamples the image by a factor of $32$ using convolutional blocks. Each block halves the resolution and increases the number of channels while utilizing inverted residual blocks \cite{mobilenetv2_cvpr2018} for efficient computation. We incorporate an additional inverted residual block for further downsampling to $1/64$ of the original size, enlarging the receptive field and enhancing the global contextual information.

\subsection{DP Encoder}
\label{subsec:dpencoder}

The DP encoder is designed to extract defocus disparity cues from DP image pairs. Exploiting the fact that DP images exhibit disparities along the epipolar line, we employ modified parallax attention \cite{pam_tpami2022} oriented along this axis to capture these cues effectively. Due to the limited disparity range (typically $-8$ to $+8$ pixels) inherent to smartphone cameras with small apertures, correspondence matching is constrained to a local neighborhood.

Formally, the left and right DP images, $I_{l}$ and $I_{r} \in \mathcal{R}^{H \times W}$, are processed by a siamese encoder comprising two shallow feature extraction modules (see Fig.~\ref{fig:network_arch}). Each module incorporates an inverted residual block \cite{mobilenetv2_cvpr2018} to mitigate spatial information loss during feature extraction---crucial given the limited spatial extent of disparity cues in DP images \cite{dp_depth_iccv2019}. Each block downsamples the input by a factor of 2. The resulting DP left and right feature maps from a block are denoted as $F_{l}$ and $F_{r} \in \mathcal{R}^{H_{f} \times W_{f} \times C}$, respectively, where $C$ represents the number of channels, with $H_{f}=\frac{H}{2}$ and $W_{f}=\frac{W}{2}$ for the first block, and $H_{f}=\frac{H}{4}$ and $W_{f}=\frac{W}{4}$ for the second. To preserve disparity information, we limit the DP encoder to two blocks, as excessive downsampling can lead to information loss and impact performance (see Section~\ref{subsec:ablationstudy}). Subsequently, the WBiPAM module processes $F_{l}$ and $F_{r}$ from each block independently to extract local defocus disparity cues. This approach effectively leverages the DP encoder to capture the rich disparity information present in DP image pairs.

\subsection{WBiPAM Module}
\label{subsec:wbipammodule}

\begin{figure}[t!]
	\centering
	\includegraphics[width=1.0\linewidth]{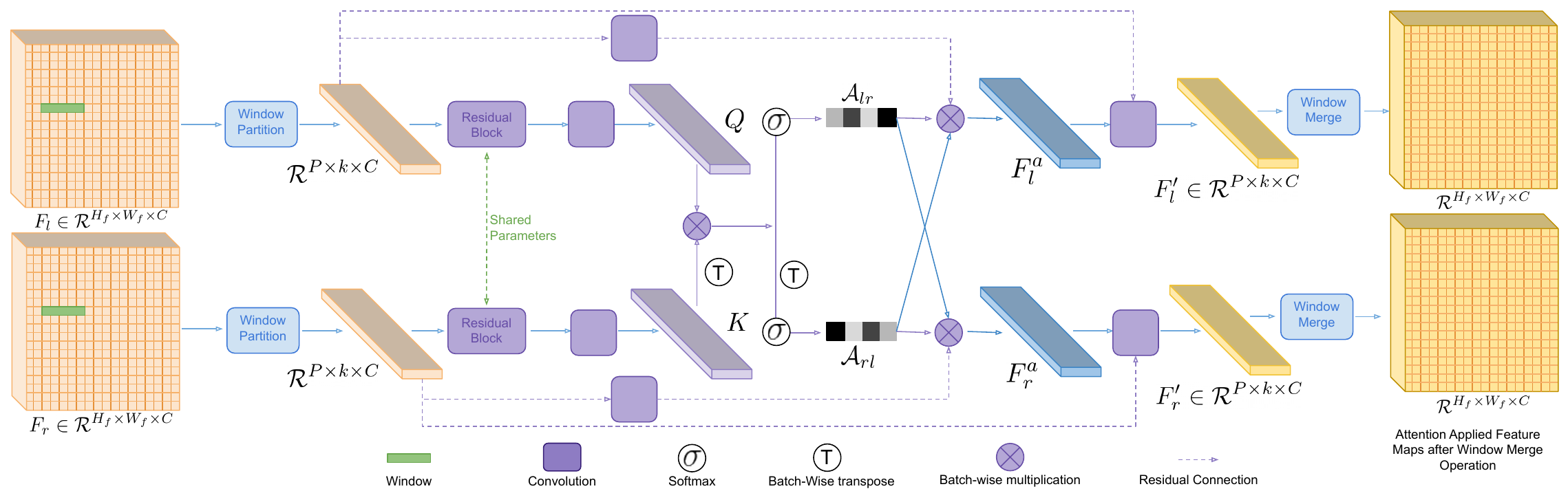}
	\vspace{-16pt}
	\caption{Design of the proposed WBiPAM module. Zoom-in using Adobe Reader.}
	\label{fig:wbipam}
	\vspace{-18pt}
\end{figure}

Fig.~\ref{fig:wbipam} shows the detailed design of the WBiPAM module. The proposed WBiPAM module offers a novel mechanism for effective disparity cue extraction from DP image pairs. It operates on DP image features, represented as $F_{l}$ and $F_{r} \in \mathcal{R}^{H_{f} \times W_{f} \times C}$. Since DP disparity is confined to a fixed range across the epipolar line, we consider a window (rectangular) within this line instead of considering the entire line. We consider a rectangular window of size $k \times 1$. To implement this, first a reshape operation is performed to divide the feature maps into $k \times k$ non-overlapping windows, restructuring $F_l$ and $F_r$ into feature maps of shape $\mathcal{R}^({\frac{H_f}{k} \times \frac{W_f}{k}) \times k \times k \times C}$. $F_l$ and $F_r$ are reformulated as feature maps of dimension $\mathcal{R}^{P \times k \times C}$ (see output of the \emph{Window Partition} blocks in Fig~\ref{fig:wbipam}), where $P = \frac{H_f}{k} \times \frac{W_f}{k} \times k$ is merged with the batch dimension. The transformed feature map of the shape $\mathcal{R}^{P \times k \times C}$, is essentially a stack of $k \times 1$ sized windows with $C$ channels.

Subsequently, a residual convolution module processes $F_l$ and $F_r$, preserving spatial and channel dimensions to retain crucial feature information. The core of the WBiPAM module is its window attention mechanism. With $F_l$ as the reference, learnable projections $W_q$ and $W_k$ generate query and key matrices from $F_l$ and $F_r$, respectively. This facilitates the computation of cross-attention scores, encapsulated in the attention score $\mathcal{A}_{lr}$ through the operation $\text{softmax}(QK^{T})$. This attention score modulates the feature map $F_l$, ultimately producing $F'_l$.

\vspace{-10pt}
\begin{subequations}
	\begin{equation}
		Q = W_q . F_l
	\end{equation}
	
	\vspace{-12pt}
	
	\begin{equation}
		K = W_k . F_r
	\end{equation}
	
	\vspace{-12pt}
	
	\begin{equation}
		\mathcal{A}_{lr} = \text{softmax}(QK^{T})
	\end{equation}
	
	\vspace{-12pt}
	
	\begin{equation}
		F^{a}_{l} = \mathcal{A}_{lr} . F_l
	\end{equation}
\end{subequations}
\vspace{-14pt}

$F^{a}_{l}$ is concatenated with the original feature map $F_l$, and subsequently processed by a convolution block, producing the final output of the WBiPAM module for the left feature map, denoted as $F'_{l} \in \mathcal{R}^{P \times k \times C}$. 

Analogously, employing the right feature map $F_r$ as the reference, an identical procedure yields the final output for the right feature map, denoted as $F'_r \in \mathcal{R}^{P \times k \times C}$.

\vspace{-10pt}
\begin{subequations}
	\begin{equation}
		\mathcal{A}_{rl} =  \mathcal{A}_{lr}^{T}
	\end{equation}
	
	\vspace{-12pt}
	
	\begin{equation}
		F^{a}_{r} = \mathcal{A}_{rl} . F_r
	\end{equation}
\end{subequations}
\vspace{-14pt} 

Subsequently, $F'_l$ and $F'_r$ undergo window merging, restructuring them into the original spatial dimensions $\mathcal{R}^{H_f \times W_f \times C}$ (see \emph{Window Merging} blocks in Fig.~\ref{fig:wbipam}). These refined feature maps are then fused with the encoding produced by the RGB encoder. As stated earlier, we limit the DP encoder to two blocks, as excessive downsampling can lead to information loss. Our empirical analysis demonstrates that the WBiPAM module significantly enhances depth estimation performance by computing attention scores based on both left and right images (see Section~\ref{subsec:ablationstudy}).

\subsection{Fusion Module}
\label{subsec:fusionmodule}

To enrich the RGB encoder's representation with local disparity cues extracted by the DP encoder, we propose a fusion mechanism that integrates $F'_l$, $F'_r$ with the corresponding RGB encoding, denoted as $F_i$. Part (b) in Fig.~\ref{fig:network_arch} demonstrates the fusion mechanism. Importantly, $F'_l$, $F'_r$, and $F_i$ share identical dimensions $\mathcal{R}^{H_{f} \times W_{f} \times C}$, where $H_{f}$, $W_{f}$, and $C$ represent the height, width, and channel count, respectively. The fusion process initiates with channel-wise concatenation of $F'_l$, $F'_r$, and $F_i$, yielding a feature map of dimensions $\mathcal{R}^{H_{f} \times W_{f} \times (3 \times C)}$. Subsequently, this concatenated representation is modulated by two convolutional layers to predict a feature-wise score $A_f \in \mathcal{R}^{H_{f} \times W_{f} \times 3}$. In $A_f$, the first, the second, and the third channel represent scaling weight for $F'_l$, $F'_r$, and $F_i$ feature maps, respectively. This feature-wise computation allows for adaptive recalibration of feature maps based on their importance for depth estimation. The computed $A_f$ is then employed to generate weighted feature maps, $F_{ilr}$, which are further processed by a convolutional module to produce the final fused feature map $F'_{ilr} \in \mathcal{R}^{H_{f} \times W_{f} \times C}$. Due to the shallow nature of the DP encoder (two layers), the fusion of DP encodings is limited to the first two RGB encoding blocks.

\subsection{Decoder}
\label{subsec:decoder}

Complementary to the encoder, the decoder progressively upsamples the feature resolution while reducing the channel count in each subsequent decoding block. Each block employs upsampling layers followed by $2D$ convolutions with PReLU activation. The final output layer utilizes a sigmoid activation to produce the depth map. To facilitate gradient flow and leverage crucial low-level information, skip connections are established between corresponding RGB encoding and decoding blocks, following the UNet \cite{unet_miccai2015} design. The feature map resolution is doubled within each decoder block, and the number of channels is halved. The decoder generates a depth map $\hat{d} \in \mathcal{R}^{H \times W \times 1}$, along with intermediate predictions from each decoding block, thus enabling deep supervision during training.

\subsection{CmTL Mechanism}
\label{subsec:cmtlmechanism}

While RGB-D datasets are abundant \cite{nyuv2_eccv2012,kitti_ijrr2013,redweb_cvpr2018,megadepth_cvpr2018,rank_loss_adobe_cvpr2020,midas_tpami2020}, RGB-DP-D datasets are scarce. However, DiFuse-Net's modality decoupled design presents a unique opportunity: leveraging existing large-scale RGB-D datasets to enhance generalization and performance. We propose a novel cross-model transfer learning (CmTL) approach comprising three stages to exploit this advantage.

\textbf{Stage 1:} The DP Encoder and Decoder subnetworks are trained end-to-end using DP-D pairs from the RGB-DP-D dataset. This stage focuses on optimizing the DP Encoder to capture the unique characteristics and challenges of depth estimation from DP images. 

\textbf{Stage 2:} The RGB Encoder and Decoder subnetworks are trained end-to-end on large-scale RGB-D datasets \cite{nyuv2_eccv2012,kitti_ijrr2013,redweb_cvpr2018,megadepth_cvpr2018,rank_loss_adobe_cvpr2020,midas_tpami2020}. Critically, training on these extensive datasets---as opposed to the smaller RGB-DP-D dataset---yields significantly improved accuracy and robustness in the RGB Encoder, capitalizing on the wealth of diverse scenes and depth variations present in these resources.

\textbf{Stage 3:} The complete DiFuse-Net is trained end-to-end, initializing the DP and RGB Encoders with weights from stages 1 and 2, respectively. The Fusion module and the Decoder weights are initialized randomly \cite{kaiming_init_iccv2015}. This knowledge transfer facilitates effective joint learning from both modalities. Empirical results (see Tab.~\ref{tab:quantitativecomparison}) demonstrate that CmTL significantly boosts RGB-DP based depth estimation.

\section{DUAL-CAMERA DUAL-PIXEL DATASET GENERATION}
\label{sec:datasetgeneration}

The Google Pixel $2/3$ dataset \cite{dp_depth_iccv2019}, while valuable, suffers from sparse and often erroneous ground-truth depth (Fig.~\ref{fig:google_vs_our_dataset}). To address this, we propose a methodology for constructing a high-quality RGB-DP-D dataset. Existing depth sensors \cite{nyuv2_eccv2012,kitti_ijrr2013} lack DP capture and pose alignment challenges.  Unlike the complex multi-camera setup in \cite{dp_depth_iccv2019}, we employ a simplified symmetric stereo configuration using two DP sensor enabled smartphones, leveraging recent AI stereo disparity estimation advancements \cite{crestereo_cvpr2022,wavelet_synth_disp_cvpr2020}. Our setup prioritizes two key criteria for high-quality data:

\begin{enumerate}
	\itemsep0em
	\item Minimal DP data processing to preserve subtle defocus disparity cues.
	\item Dense, accurate ground-truth depth maps aligned with RGB-DP images.
\end{enumerate}

\subsection{Symmetric Stereo Camera Setup}
\label{subsec:camerasetup}

\begin{figure}[t!]
	\centering
	\includegraphics[width=0.98\linewidth]{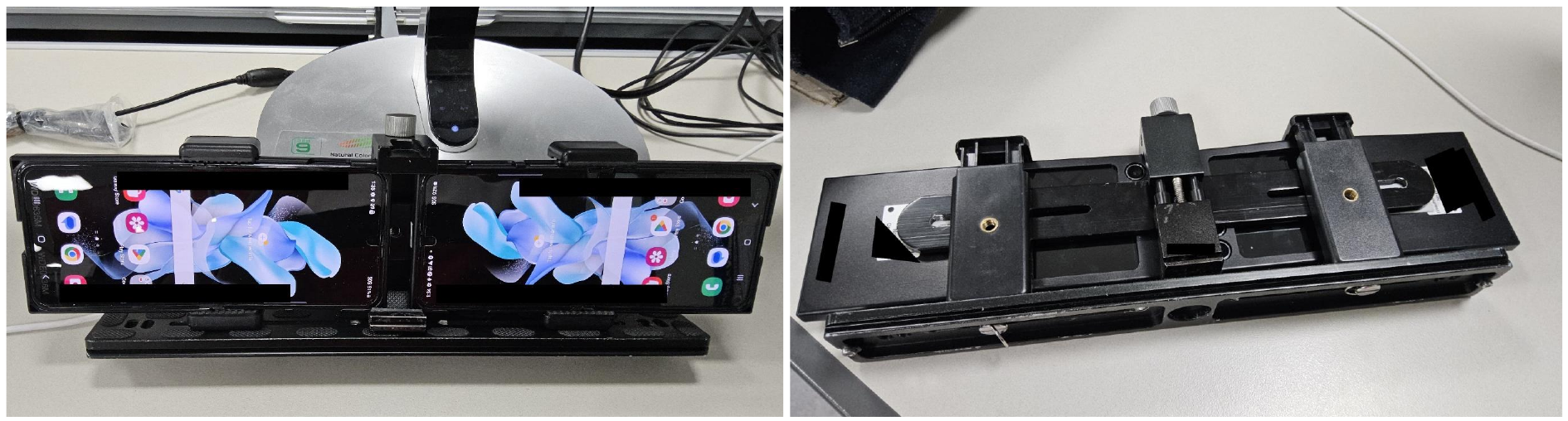}
	\vspace{-8pt}
	\caption{Our symmetric stereo camera setup for data acquisition.}
	\label{fig:symmcamsetup}
	\vspace{-8pt}
\end{figure}

Symmetric stereo cameras simplify calibration and rectification compared to wide/telephoto combinations \cite{taxonomy_schar_ijcv_2002}. We employed two Samsung Galaxy S$23$ Ultra smartphones in a symmetric stereo configuration (Fig.~\ref{fig:symmcamsetup}), using the front cameras to minimize baseline ($2.5 cm$) and occlusions. Rigid camera holders and a metal support ensured stability. Simultaneous capture was achieved using Galaxy S-pens and USB-C camera switch.

\subsection{Stereo Rectification}
\label{subsec:rectification}

\begin{figure}[t!]
	\centering
	\includegraphics[width=0.98\linewidth]{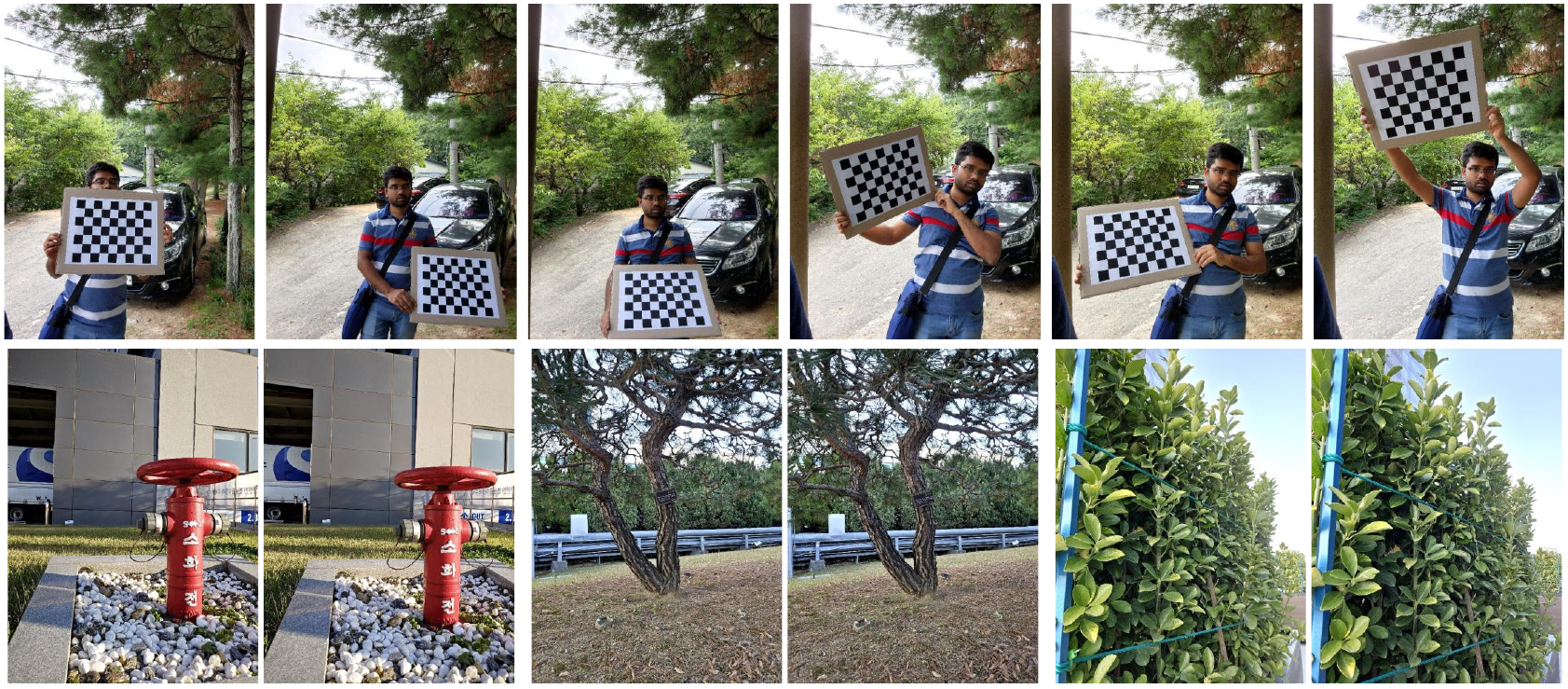}
	\vspace{-8pt}
	\caption{Top row: Checkerboard images captured in the begging of a capture session. Bottom row: Sample stereo captures.}
	\label{fig:calib_and_samples}
	\vspace{-16pt}
\end{figure}

Our camera setup primarily exhibits horizontal disparity. To ensure precise alignment, we perform stereo calibration and rectification before each capture session, utilizing $30$-$40$ checkerboard samples. Each session yields $120$-$150$ images (Fig.~\ref{fig:calib_and_samples}). Such a routine protocol mitigates potential positional shifts in long capture sessions. Calibration and rectification are performed at half-resolution, with the resulting coordinate mapping applied to full-resolution images. Rectified stereo pairs are processed through an AI stereo disparity estimation pipeline.

\begin{figure}[t!]
	\centering
	\includegraphics[width=0.98\linewidth]{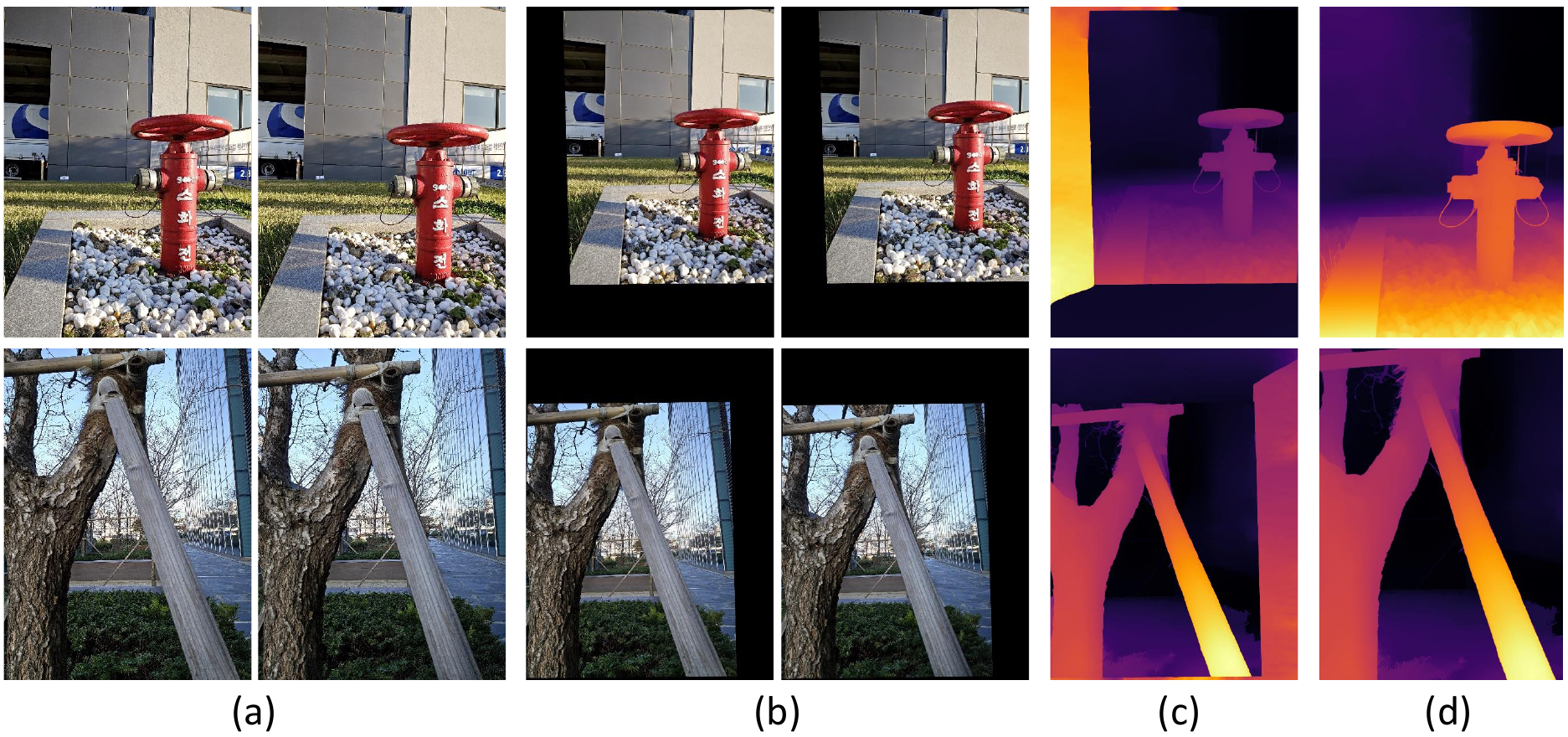}
	\vspace{-8pt}
	\caption{(a) Stereo pairs. (b) Rectified stereo pairs. (c) Estimated disparity map in the rectified plane. (d) Disparity map projected back to the original plane along with $40$-pixel border crop (the same crop is later applied to original RGB-DP images for alignment).}
	\label{fig:data_gen_samples}
	\vspace{-8pt}
\end{figure}

To address the misalignment between the estimated disparity map (in the rectified plane) and the captured DP images, we project the disparity map back to the original plane, thus preserving DP cues. This is preferable to rectifying DP images, which would distort DP cues. Although reversing distortion is less precise than inverting pose parameters, our minimal distortion coefficients allow us to approximate reverse distortion by inverting their signs. A $40$-pixel border crop mitigates any residual inaccuracies (Fig.~\ref{fig:data_gen_samples}).

\newcommand{\dbgencheckerboardsamplesfigwidth}[0]{0.15\linewidth}

\newcommand{\dbgensamplesfigwidth}[0]{0.14\linewidth}

\subsection{Ground-truth Depth Generation}
\label{subsec:gtdepthgeneration}

We leverage advances in AI stereo disparity estimation \cite{crestereo_cvpr2022,wavelet_synth_disp_cvpr2020} to generate accurate disparity maps from rectified stereo pairs. Our model, trained on a large synthetic dataset \cite{sceneflow_dataset_cvpr2016} augmented with slight vertical distortions, ensures robustness to minor rectification errors. Back-projecting the estimated disparity map and applying a $40$-pixel border crop yields aligned RGB-DP-D samples with preserved DP cues.

\begin{figure}[t!]
	\centering
	\includegraphics[width=0.98\linewidth]{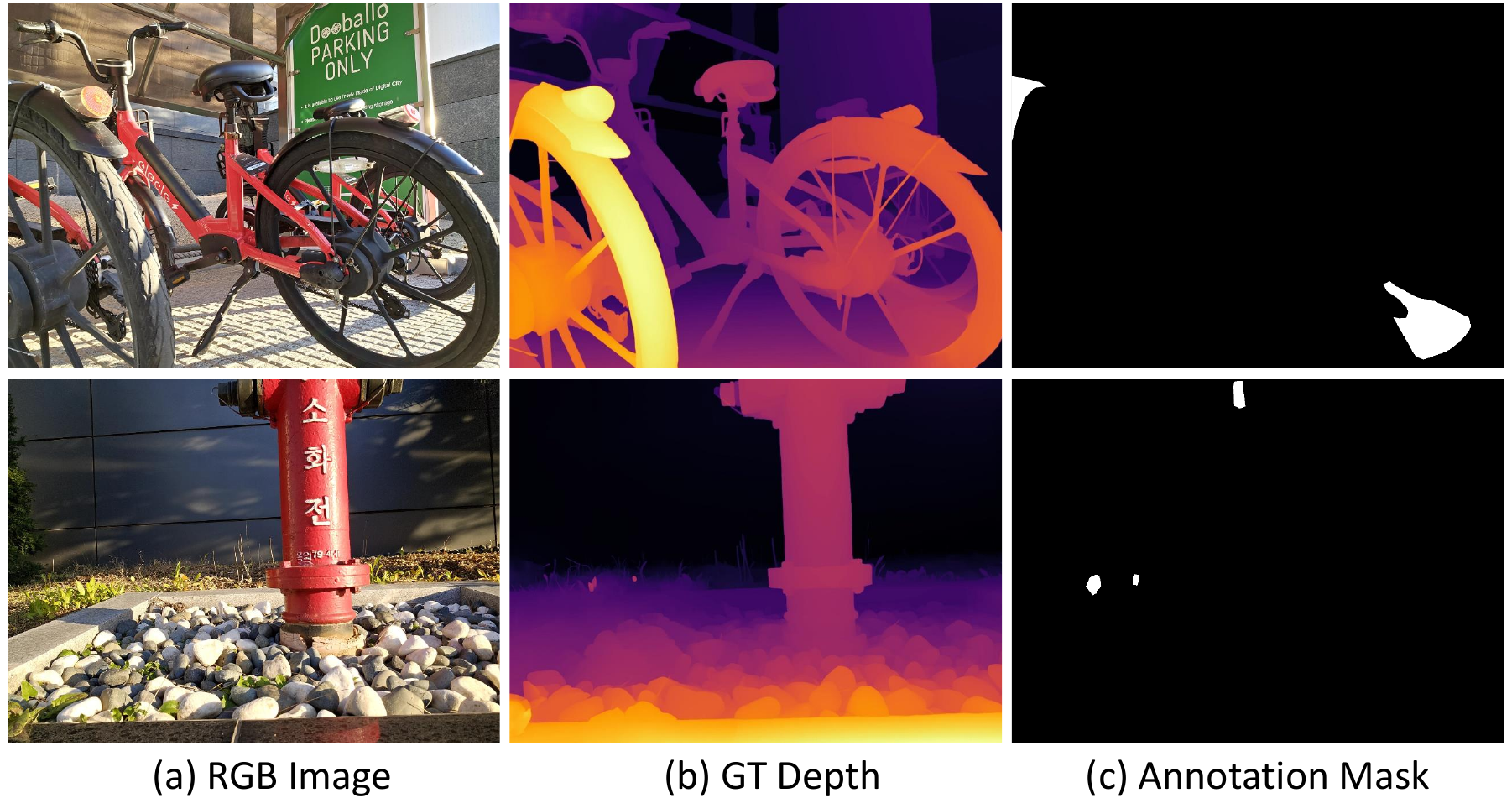}
	\vspace{-8pt}
	\caption{Example annotations from our manual quality control process. The binary masks indicate regions of significant error in the ground-truth disparity map, identified through visual inspection. The masked regions are excluded from loss computation during training.}
	\label{fig:annotationsamples}
	\vspace{-8pt}
\end{figure}

To ensure dataset quality, we employ manual annotation (Fig.~\ref{fig:annotationsamples}) to mask visually incorrect disparity estimations. This meticulous process guarantees high-quality ground truth, with masked regions excluded from loss computation during training.

\begin{figure}[t!]
	\centering
	\includegraphics[width=1.0\linewidth]{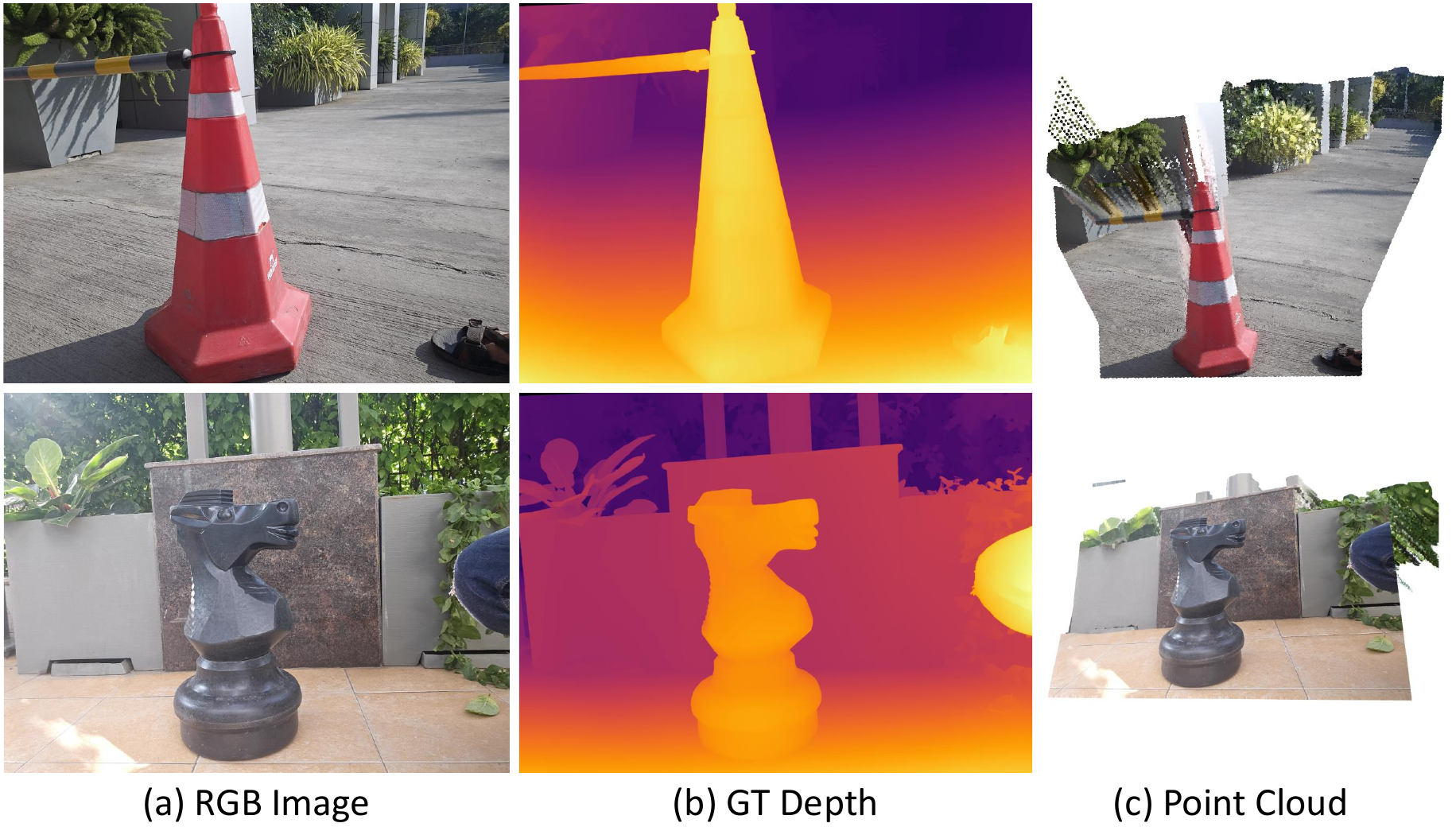}
	\vspace{-20pt}
	\caption{Qualitative assessment of the ground-truth depth accuracy in the DCDP dataset. Point clouds are rendered from a novel viewpoint to highlight the geometric consistency in the reconstructed $3$D structure.}
	\label{fig:dcdppointclouds}
	\vspace{-8pt}
\end{figure}

Finally, Fig.~\ref{fig:dcdppointclouds} provides a qualitative assessment of the ground-truth depth accuracy in the DCDP dataset. By rendering the point clouds from a novel viewpoint, we highlight the geometric consistency and absence of artifacts in the reconstructed $3$D structure.

\section{EXPERIMENTS}
\label{sec:experiments}

Our experimental evaluation comprises two primary components: (i) assessment and benchmarking of the proposed DiFuse-Net method, and (ii) evaluation on our new high-quality DCDP dataset.  For a fair comparison with prior work \cite{dp_depth_iccv2019}, we initially train and evaluate DiFuse-Net on the publicly available Google DP dataset \cite{dp_depth_iccv2019}. Additionally, we conduct an ablation study on the same dataset to analyze the contribution of individual components within DiFuse-Net. Finally, we present results showcasing the advantages of utilizing our DCDP dataset for training, along with benchmarking results on the DCDP test set.

\subsection{Datasets}
\label{subsec:datasets}

\textbf{Google DP Dataset:} Google DP dataset \cite{dp_depth_iccv2019} consists of $12,530$ training and $684$ test RGB-DP-D samples. Since each scene is captured by $5$ cameras in the multi-camera rig, effectively there are $2506$ novel images in this dataset.

\textbf{DCDP:} Our DCDP dataset consists of $5000$ novel training and $700$ novel test RGB-DP-D samples.

\subsection{Implementation}
\label{subsec:implementation}

We used the PyTorch \cite{pytorch} framework for implementation. For training, we used Adam optimizer \cite{adam_opt_iclr2015} with momentum of $0.9$. We employed a polynomial learning rate decay scheduler with power term set to $0.9$, and an initial learning rate of $1 \times 10^{-4}$. Similar to the loss function in \cite{midas_tpami2020}, we combine the scale-invariant mean absolute error with a scale-invariant gradient matching term. The gradient matching term helps to preserve sharp edges and boundaries \cite{midas_tpami2020}. Also, similar to \cite{dp_depth_iccv2019} the model is trained to predict the inverse depth.

\vspace{-8pt}
\begin{equation} 
	\begin{array}{l}
		\mathcal{L} = MAE(d, \hat{d}) + \lambda\times Grad(d, \hat{d})
		
	\end{array}
	\label{eqn:loss}
\end{equation}
\vspace{-8pt}

In Eq.~\ref{eqn:loss}, $d$ signifies the ground-truth inverse depth, whereas $\hat{d}$ denotes the predicted affine invariant depth. $\lambda$ is a hyper-parameter, set to $30$ across all experiments. MAE($\cdot$) and Grad($\cdot$) represent the mean absolute error and the gradient loss functions, respectively.

\subsection{Evaluation Metrics}
\label{subsec:evaluationmetrics}

Similar to \cite{dp_depth_iccv2019}, we use Spearman's Rank Correlation Coefficient (SRCC) to evaluate the ordinal correctness of the estimated depth map with ground-truth depth, and affine-invariant versions of mean absolute error (MAE) and root mean squared error (RMSE), denoted by AIWE~$1$ and AIWE~$2$ respectively.

\section{Results and Discussions}
\label{sec:resultsanddiscussion}

\subsection{Evaluation and Comparison}
\label{subsec:evalcomparison}

Tab.~\ref{tab:quantitativecomparison} shows detailed quantitative comparison of DiFuse-Net with other baselines, viz., \textbf{DPNet} \cite{dp_depth_iccv2019}, \textbf{Baseline}, and \textbf{Stereo Baseline} \cite{raft_stereo_3dv2021}. We report the latest DPNet metrics provided on their \href{https://github.com/google-research/google-research/blob/master/dual_pixels/README.md#results-and-evaluation}{official GitHub page} due to their release of modified train and test datasets. This ensures a fair evaluation of our method against DPNet. Baseline is the enhanced DPNet model by increasing the convolutional layers, aligning the total number of parameters with that of DiFuse-Net, to ensure a more fair comparison. Stereo Baseline replaces WBiPAM with a traditional stereo matching cost-volume approach \cite{raft_stereo_3dv2021,wavelet_synth_disp_cvpr2020,taxonomy_schar_ijcv_2002}, providing a comparison with established stereo methods. More specifically, we use \cite{raft_stereo_3dv2021}. The total number of parameters of Stereo Baseline are aligned with that of DiFuse-Net to ensure a fair comparison.

As it can be seen in Tab.~\ref{tab:quantitativecomparison}, the DiFuse-Net model outperforms other models in all the metrics, demonstrating the efficacy of our approach in disentangling RGB and DP images, WBiPAM module and CmTL mechanism.

\begin{figure*}[t!]
	\includegraphics[width=1.0\linewidth]{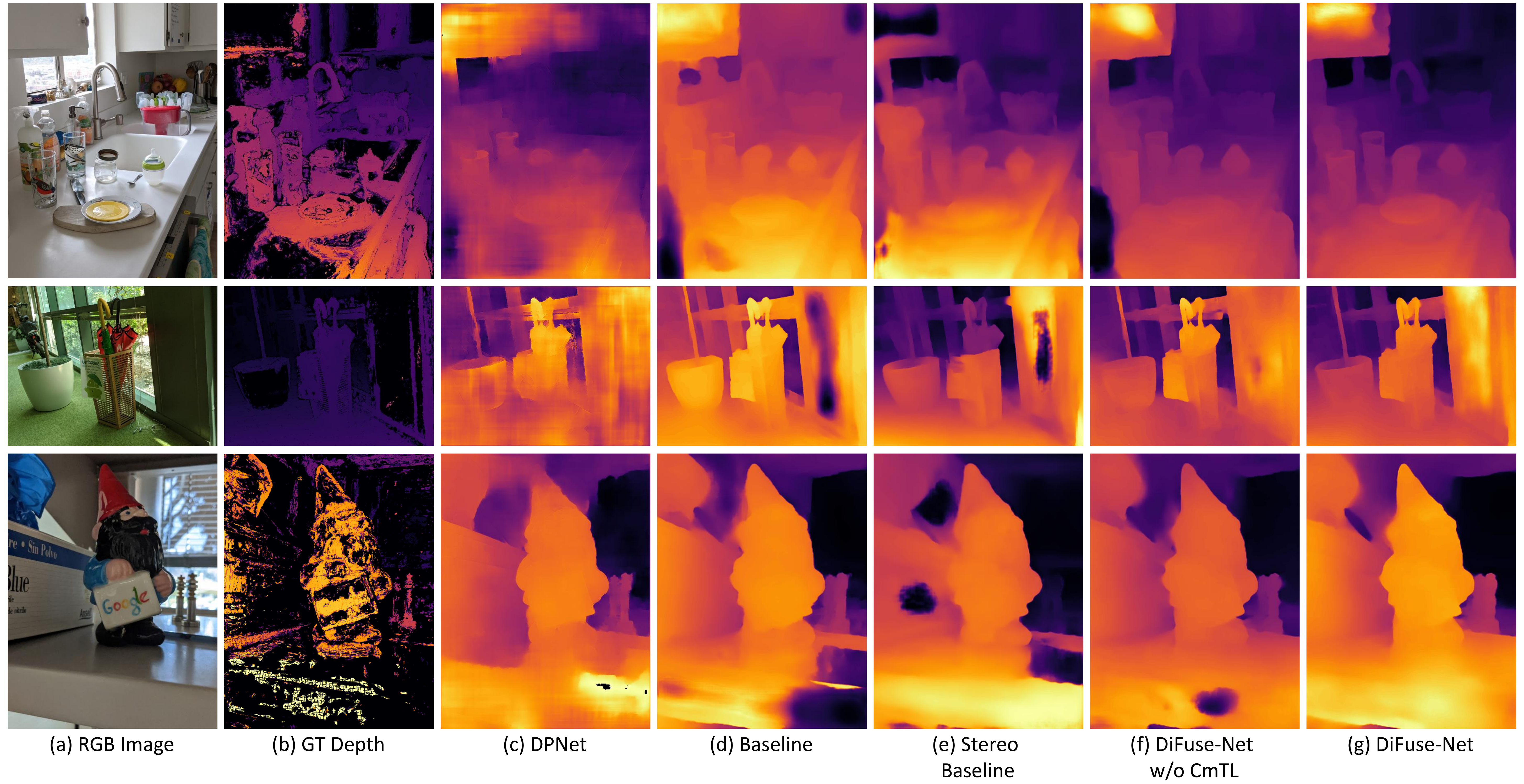}
	\vspace{-20pt}
	\caption{Qualitative evaluation of DiFuse-Net with baseline methods, viz., DPNet \cite{dp_depth_iccv2019}, Baseline, Stereo Baseline \cite{raft_stereo_3dv2021}, and DiFuse-Net w/o CmTL on Google DP dataset.}
	\label{fig:Qualitative}
	\vspace{-12pt}
\end{figure*}

Fig.~\ref{fig:Qualitative} presents the qualitative comparison between DPNet, Baseline, Stereo Baseline, DiFuse-Net w/o CmTL and DiFuse-Net methods. As previously discussed, in images containing texture-less regions, DP images often lack discernible disparity, resulting in poor depth output within these regions. However, our approach of decoupling RGB and DP images enables the model to extract global information more effectively from RGB images, consequently mitigating the instances of poor depth output substantially.

\subsection{Ablation Study}
\label{subsec:ablationstudy}

In our ablation study, we systematically investigate the design choices in the WBiPAM module, Fusion module, and the DP Encoder network depth. Tab.~\ref{tab:ablation_study} presents the quantitative results.  To ensure a fair comparison, the number of parameters in the remaining network components is adjusted for each ablative variant.

First, we ablate the WBiPAM module as follows.
\begin{itemize}
	\itemsep0em
	\item \textbf{No WBiPAM:} This variant removes the WBiPAM module entirely, highlighting its contribution to performance.
	\item \textbf{No Window WBiPAM:} Here, we remove the window partitioning within WBiPAM, allowing cross-attention to operate across the entire feature map, thus evaluating the importance of local context.
	\item \textbf{Unidirectional WBiPAM:}  This variant uses only the left DP features $F_l$ as input to WBiPAM, assessing the benefits of bidirectional attention.
\end{itemize}

The results in Tab.~\ref{tab:ablation_study} clearly demonstrate that WBiPAM, with its windowed bidirectional attention mechanism, outperforms all other variants.

Next, our exploration of fusion strategies reveals that feature-wise fusion of modalities (as explained in Sec.~\ref{subsec:fusionmodule}) is superior to more granular pixel-wise (recalibration of each pixel) or channel-wise (recalibration of each channel) fusion.

Regarding the DP Encoder, a two-layer architecture proves optimal. We hypothesize that additional downsampling layers lead to the loss of subtle defocus disparity cues, hindering performance. 

Finally, ablation study on the CmTL mechanism shows that the DiFuse-Net model additionally pre-trained on large-scale RGB-D datasets using our CmTL mechanism obtains the superior performance.

\begin{table}[t!]
	\centering
	\caption{Quantitative comparison of DiFuse-Net on Google DP dataset.}
	\vspace{-8pt}
	\label{tab:quantitativecomparison}
	{
		\renewcommand{\arraystretch}{0.87}
		\setlength{\tabcolsep}{0.5em}
		\resizebox{0.4\textwidth}{!}{
			\begin{tabular}{lccccc}\\
				\toprule
				\textbf{Model} & 1 - SRCC $\; \downarrow$  & AIWE 1 $\; \downarrow$ & AIWE 2 $\; \downarrow$\\
				\midrule
				DPNet\cite{dp_depth_iccv2019} & $0.1520$ & $0.0181$  & $0.0268$\\
				Baseline & $0.0927$ & $0.0142$  & $0.0218$\\
				Stereo Baseline \cite{raft_stereo_3dv2021} & $0.0911$ & $0.0137$ &$0.0216$\\
				DiFuse-Net & $\textbf{0.0799}$ & $\textbf{0.0128}$ & $\textbf{0.0202}$\\
				\bottomrule
			\end{tabular}
		}
	}
	\vspace{-8pt}
\end{table}

\begin{table}[!t]
	\centering
	\caption{Quantitative results of our ablation study on Google DP dataset.}
	\vspace{-8pt} 
	\label{tab:ablation_study}
	{	
		\renewcommand{\arraystretch}{0.90}
		\setlength{\tabcolsep}{0.5em}
		\resizebox{0.48\textwidth}{!}{
			\begin{tabular}{l|c|c|c|c}
				\toprule
				Pipeline Component & Ablation Details  & 1 - SRCC$\downarrow$   & AIWE 1$\downarrow$  &  AIWE 2$\downarrow$ \\
				\midrule
				\multirow{ 4}{*}{WBiPAM} & No WBiPAM & $0.0865$ & $0.0132$ & $0.0210$ \\
				& No Window WBiPAM & $0.0912$ & $0.0137$ & $0.0214$\\
				& Unidirectional WBiPAM& $0.1454$ & $0.0191$ & $0.0276$\\ 
				
				\midrule
				\multirow{ 3}{*}{Fusion Module}
				& Pixel-wise Fusion & $0.0919$ & $0.0138$ & $0.0214$ \\
				& Channel-wise Fusion & $0.0855$ & $0.0129$ & $0.0204$\\ 
				
				\midrule
				\multirow{ 4}{*}{DP Encoder}
				& 5 Layer DP Encoder & $0.0950$ & $0.0138$ & $0.0215$ \\
				& 4 Layer DP Encoder & $0.0889$ & $0.0135$ & $0.0210$ \\
				& 3 Layer DP Encoder & $0.0868$ & $0.0134$ & $0.0210$ \\ 
				& 1 Layer DP Encoder & $0.0859$ & $0.0135$ & $0.0210$ \\ 
				
				\midrule
				\multirow{ 1}{*}{CmTL}
				& DiFuse-Net w/o CmTL & $0.0833$ & $0.0129$ & $0.0205$ \\
				
				\midrule
				\multicolumn{2}{c|}{DiFuse-Net} & $\textbf{0.0799}$ & $\textbf{0.0128}$ & $\textbf{0.0202}$ \\
				\bottomrule
			\end{tabular}
		}
	}
	\vspace{-4pt}
\end{table}

\begin{table}[t!]
	\centering
	\caption{Additional comparison of DiFuse-Net with monocular depth estimation methods trained on large and diverse RGB-D datasets, Google DP test dataset is used for this evaluation.}
	\vspace{-8pt}
	\label{tab:midas_comparison}
	\resizebox{0.95\linewidth}{!}
	{
		\begin{tabular}{r|ccc}
			\toprule
			Model & 1 - SRCC $\; \downarrow$  & AIWE 1 $\; \downarrow$ & AIWE 2 $\; \downarrow$ \\
			
			\midrule
			
			MiDaS v$3.1$ $\text{BEiT}_{\text{L}-512}$ \cite{midas_tpami2020} & $0.0971$ & $0.0168$  & $0.0267$ \\
			ZoeDepth \cite{zeo_depth_arxiv2023} & $0.1168$ & $0.0272$  & $0.0379$ \\
			DiFuse-Net & $\textbf{0.0799}$ & $\textbf{0.0128}$ & $\textbf{0.0202}$ \\
			
			\bottomrule
		\end{tabular}
	}
	\vspace{-12pt}
\end{table}

\newcommand{\qualcomparisonfigwidth}[0]{0.13\linewidth}

\subsection{Additional Evaluation}
\label{subsec:additionalevaluation}

\begin{figure}[t!]
	\includegraphics[width=1.0\linewidth]{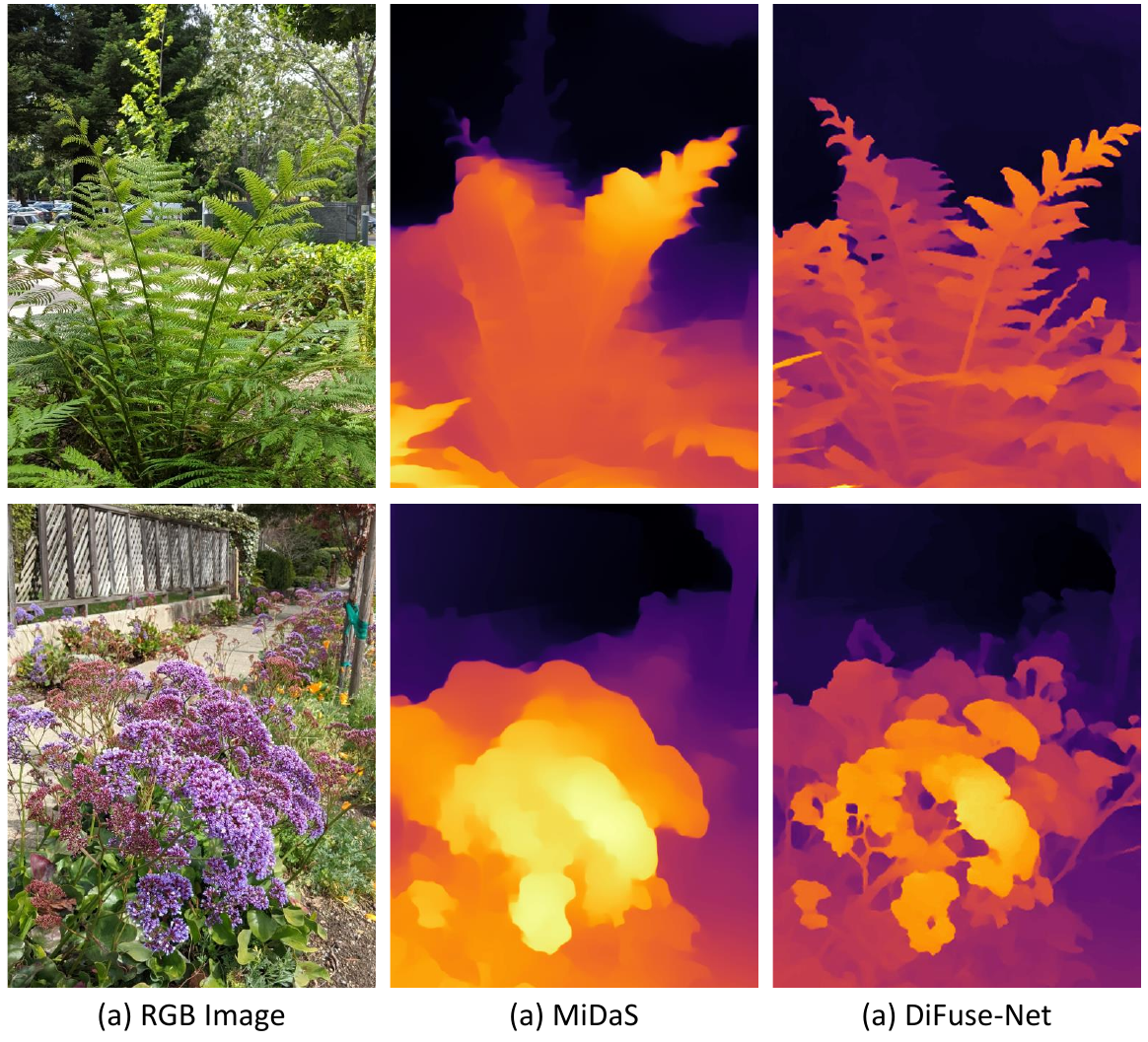}
	\vspace{-20pt}
	\caption{Qualitative comparison of DiFuse-Net with MiDaS.}
	\label{fig:midas_fig}
	\vspace{-7pt}
\end{figure}

We conducted a comparative analysis between DiFuse-Net and monocular depth estimation models trained on large-scale RGB-D datasets, viz., MiDaS v$3.1$ $\text{BEiT}_{\text{L}-512}$ \cite{midas_tpami2020} which comprises of $345$M parameters, and ZoeDepth \cite{zeo_depth_arxiv2023}. ZoeDepth uses MiDaS as backbone, and uses a metric depth prediction head. Best numbers were obtained by the ZoeD-M$12$-N (trained on NYU V$2$) variant. Our analysis revealed that despite DiFuse-Net having substantially fewer paramaters ($9.9$M), its depth map quality was better than that of MiDaS, owing to the effective use of DP images in an improved model architecture. Table ~\ref{tab:midas_comparison} shows the quantitative score comparison of DiFuse-Net with MiDaS, and ZoeDepth indicating better scores for our model. The qualitative comparisons are presented in ~\ref{fig:midas_fig}. where our model exhibits enhanced details.

\subsection{Evaluation on the DCDP Dataset}
\label{subsec:evalondcdpdataset}

\begin{figure}[t!]
	\centering
	\includegraphics[width=0.99\linewidth]{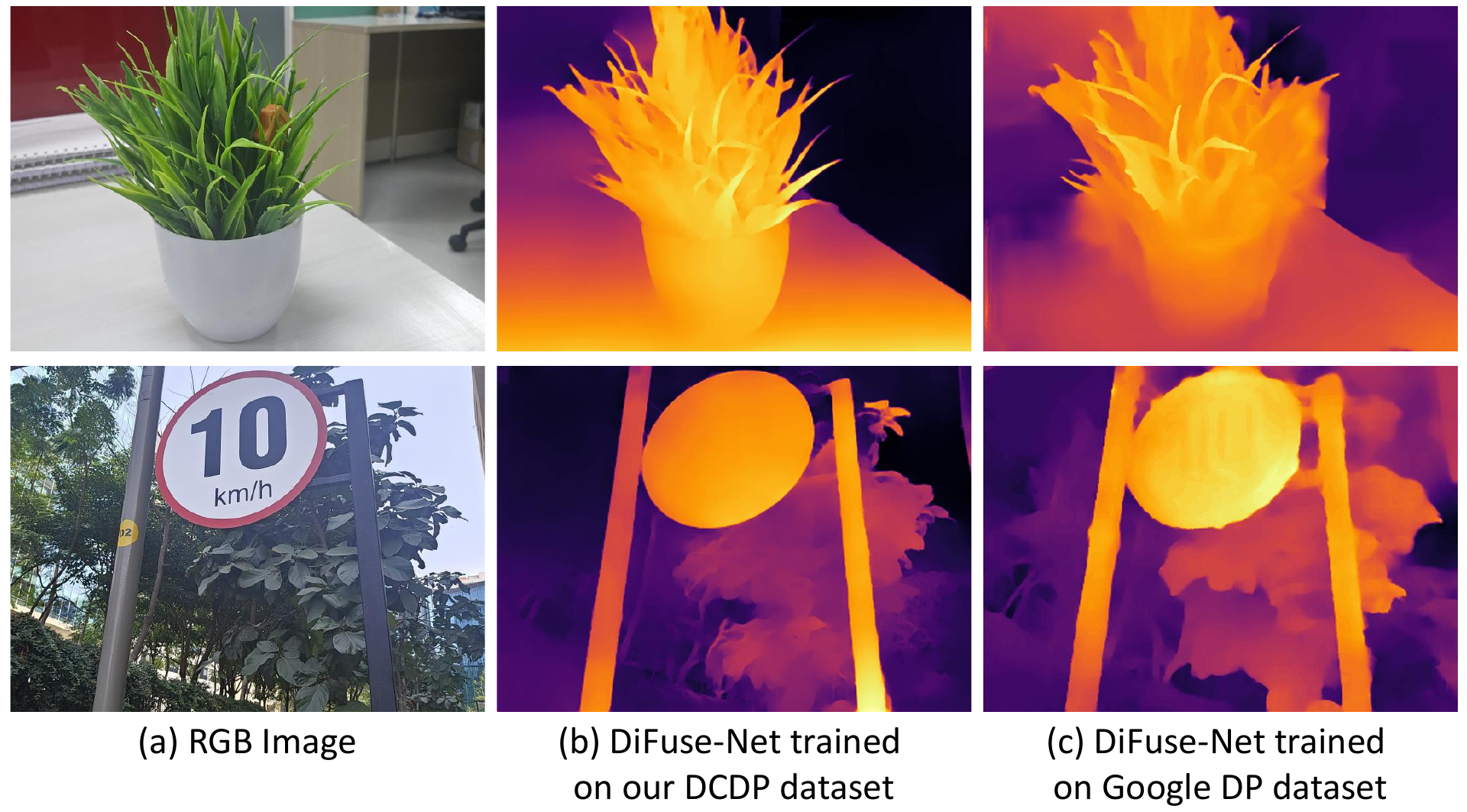}
	\vspace{-18pt}
	\caption{DCDP trained (first) vs DPNet trained (second) output. Qualitative comparison of DiFuse-Net trained on DCDP vs Google DP dataset to showcase the advantages of our DCDP dataset. It can be observed that the results in (b) exhibit depth leakage at object boundaries, with gaps erroneously filled. In contrast, the DCDP trained model consistently demonstrates sharper boundaries, preserves fine-grained gaps, and exhibits superior detail in thin structures, underscoring the benefits of our high-quality dataset.} 
	\label{fig:our_data_improvement}
	\vspace{-4pt}
\end{figure}

We report the benefit of using our DCDP dataset for training a RGB-DP based depth estimation model. Fig.~\ref{fig:our_data_improvement} shows that a DCDP trained model produces sharper and smoother depth maps due to our dense, high-quality ground-truth. Tab.~\ref{tab:dcdp_benchmarking} additionally provides the quantitative benchmarking on our new DCDP dataset.

\newcommand{\googoursdbfigwidthh}[0]{0.28\linewidth}

\begin{table}[t!]
	\footnotesize
	\centering
	\caption{Benchmarking on our DCDP dataset.} 
	\vspace{-4pt}
	\label{tab:dcdp_benchmarking}
	\resizebox{7.5cm}{!}
	{
		\centering
		\begin{tabular}{r|ccc}
			\toprule
			Model & 1 - SRCC $\; \downarrow$  & AIWE 1 $\; \downarrow$ & AIWE 2 $\; \downarrow$ \\
			
			\midrule
			
			DPNet \cite{dp_depth_iccv2019} & $0.1522$ & $0.0087$  & $0.0128$ \\
			Baseline & $0.0928$ & $0.0068$ & $0.0104$\\
			Stereo Baseline \cite{raft_stereo_3dv2021} & $0.0912$ & $0.0066$ & $0.0098$\\
			DiFuse-Net & $\textbf{0.0878}$ & $\textbf{0.0062}$ & $\textbf{0.0092}$\\
			
			\bottomrule
		\end{tabular}
	}
	\vspace{-6pt}
\end{table}

\section{CONCLUSION}
\label{sec:conclusion}

In this work, we have introduced a novel depth estimation method that harnesses the potential of ubiquitous DP sensors for extracting valuable defocus disparity cues. Our proposed DiFuse-Net architecture, with its modality-decoupled design and the innovative WBiPAM module, effectively leverages both DP and RGB information to significantly enhance depth estimation accuracy. Furthermore, our CmTL strategy overcomes the scarcity of RGB-DP-D datasets by leveraging existing large-scale RGB-D datasets, further boosting performance. Extensive evaluations demonstrate the superiority of our method over the baseline methods. Additionally, we contribute a new high-quality RGB-DP-D dataset, named DCDP, constructed using a carefully designed symmetric stereo camera setup and advanced AI stereo disparity estimation techniques. This dataset will serve as a valuable resource for future research in this domain. Our work paves the way for exploiting the full potential of DP sensors for accurate and robust depth estimation, opening doors for various applications in mobile robotics and beyond.










\bibliographystyle{IEEEtran}
\bibliography{references}

\end{document}